\documentclass[10pt, conference, compsocconf]{IEEEtran}
\ifCLASSINFOpdf
  \usepackage[pdftex]{graphicx}
\else
\fi

\usepackage{booktabs} % To thicken table lines

\usepackage{amsmath}

\usepackage{url}
\begin{document}

\title{Clarifying Myths About the Relationship Between Shape Bias, Accuracy, and Robustness}


% \author{\IEEEauthorblockN{Zahra Golpayegani}
% \IEEEauthorblockA{Gina Cody School of Engineering\\
% Concordia University\\
% Montreal, Canada\\
% zahra.golpayegani@mail.concordia.ca}
% \and
% \IEEEauthorblockN{Patrick St-Amant}
% \IEEEauthorblockA{Zetane Systems Inc.\\
% Montreal, Canada\\
% patrick@zetane.com}
% \and
% \IEEEauthorblockN{Nizar Bouguila}
% \IEEEauthorblockA{Gina Cody School of Engineering\\
% Concordia University\\
% Montreal, Canada\\
% nizar.bouguila@concordia.ca}
% }


\author{\IEEEauthorblockN{Zahra Golpayegani\IEEEauthorrefmark{1},
Patrick St-Amant\IEEEauthorrefmark{2},
Nizar Bouguila\IEEEauthorrefmark{1}
}
\IEEEauthorblockA{\IEEEauthorrefmark{1}Gina Cody School of Engineering,
Concordia University, Montreal, Canada\\zahra.golpayegani@mail.concordia.ca\\nizar.bouguila@concordia.ca}
\IEEEauthorblockA{\IEEEauthorrefmark{2}Zetane Systems Inc.\\patrick@zetane.com}
}


\IEEEoverridecommandlockouts
\IEEEpubid{\makebox[\columnwidth]{978-1-5386-5541-2/18/\$31.00~\copyright2018 IEEE \hfill}
\hspace{\columnsep}\makebox[\columnwidth]{ }}

\maketitle
\IEEEpubidadjcol

\begin{abstract}
Deep learning models can perform well when evaluated on images from the same distribution as the training set. However, applying small perturbations in the forms of noise, artifacts, occlusions, blurring, etc. to a model's input image and feeding the model with out-of-distribution (OOD) data can significantly drop the model's accuracy, making it not applicable to real-world scenarios. 
Data augmentation is one of the well-practiced methods to improve model robustness against OOD data; however, examining which augmentation type to choose and how it affects the OOD robustness remains understudied. 
There is a growing belief that augmenting datasets using data augmentations that improve a model's bias to shape-based features rather than texture-based features results in increased OOD robustness for Convolutional Neural Networks trained on the ImageNet-1K dataset. This is usually stated as ``an increase in the model's shape bias results in an increase in its OOD robustness". Based on this hypothesis, some works in the literature aim to find augmentations with higher effects on model shape bias and use those for data augmentation. By evaluating 39 types of data augmentations on a widely used OOD dataset, we demonstrate the impact of each data augmentation on the model's robustness to OOD data and further show that the mentioned hypothesis is not true; an increase in shape bias does not necessarily result in higher OOD robustness. By analyzing the results, we also find some biases in the ImageNet-1K dataset that can easily be reduced using proper data augmentation. Our evaluation results further show that there is not necessarily a trade-off between in-domain accuracy and OOD robustness, and choosing the proper augmentations can help increase both in-domain accuracy and OOD robustness simultaneously.
\end{abstract}


% \fi
\IEEEpeerreviewmaketitle

\begin{figure*}[h!]
\centering
\includegraphics[width=\textwidth]{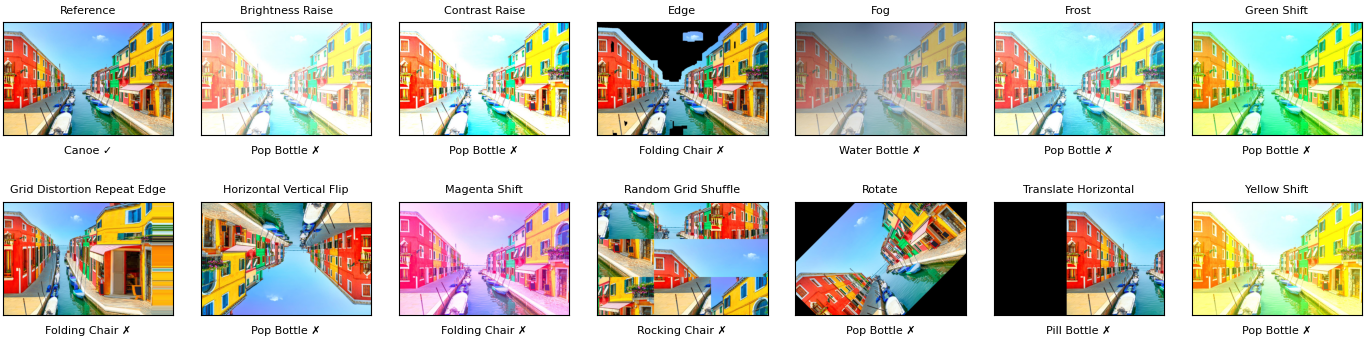}
\caption{This illustration shows how a ResNet-50 model with 89.06\% in-domain accuracy fails when a small perturbation has been applied to the image. The label on top of each image is the name of the perturbation, and the label below each image is the predicted class. "reference" is the clean image.}
\label{img:fool}
\end{figure*}

\section{Introduction}

While computer vision models have been applied to different areas of research to this date, developing models that can perform well when perturbations are present in the input image is still a challenging quest. To use computer vision in practice, we need reliable models that are robust against major and minor image perturbations. Nevertheless, even the most commonly used models fail to perform well when minor perturbations in light, color, etc. are present in the input image \cite{borkar2019deepcorrect, ghosh2018robustness, zhou2017classification, roy2018effects}. Fig. \ref{img:fool} illustrates some examples that can easily fool a trained ResNet-50 \cite{he2016deep} model with 89.06\% in-domain accuracy. To push the boundaries of computer vision, more work still needs to be done to make models reliable enough to be used even in use cases, such as in autonomous vehicles, where a faulty decision can lead to disastrous consequences.

The image perturbations can even be undetectable to a human observer and yet fool the model. A well-studied instance of this phenomenon is adversarial examples \cite{szegedy2013intriguing} which are inputs that are intentionally designed to fool the model and are indistinguishable to human eyes. Different methods have been proposed to make models robust against adversarial attacks \cite{chen2020adversarial, kurakin2016adversarial, bashivan2021adversarial, herrmann2022pyramid, wang2019direct, tzeng2017adversarial}. In addition to the adversarial examples, other changes can be applied to images, especially perturbations that are more common in nature, such as changes in light, color shifts, and different weather conditions. These types of image perturbations are the main focus of study in this work since they are present in many real-world applications.

Data augmentation is a well-practiced method to improve model robustness. Some data augmentation libraries already exist from previous research work, including imgaug \cite{imgaug}, Albumentations \cite{buslaev2020albumentations}, AugLy \cite{papakipos2022augly}, and torchvision \cite{torchvision}. In none of the above frameworks, however, authors propose a concrete solution based on which users can find which augmentations are best to use to improve their model performance. Some previous works focus on finding the optimal augmentation automatically. For instance, in \cite{cubuk2018autoaugment}, a reinforcement learning-based framework has been proposed that searches for the best augmentation policies in a discrete space. However, because of the high complexity of the search space, when the number of augmentations that are required to be investigated is more than a few, especially in cases where the resources are limited, this method cannot be used at all. Developing a fast and efficient solution to find which augmentations fit a specific learning problem is crucial as currently many data scientists rely on trial and error to choose an augmentation which requires a lot of resources.

Based on the experiments in \cite{geirhos2018imagenet}, it is hypothesized that by adding more shape bias to a network, model robustness will be improved. While some works are based on the correctness of this hypothesis \cite{lee2022improving}, there is not any extensive study in the literature that examines this claim in practice. However, there is a growing belief that using augmentations that add shape bias to the network will improve model robustness. 

% created a stylized version of the ImageNet \cite{deng2009imagenet}  dataset, called Stylized ImageNet (SIN), and proved that a SIN-trained ResNet outperforms an ImageNet-trained ResNet on most of the image perturbations considered in their work. According to this result, it seems like since SIN-trained ResNets learn a shape-based representation of ImageNet, they are more robust than ImageNet-trained ResNets that mostly focus on texture information. Therefore, the authors claim that ImageNet data induce texture bias in a network, and they hypothesize that by adding more shape bias to a network, we can improve model robustness. 

In this paper, by evaluating a diverse set of augmentations, including simulating weather conditions, occlusions, noise, color changes, etc., we study how each data augmentation function affects the model's robustness to out-of-distribution (OOD) data. Furthermore, we use a shape bias metric introduced in \cite{geirhos2021partial} to measure the shape bias imposed by each augmentation function on the trained model, and we conclude that although in some cases there is a correlation between the model's shape bias and its robustness to OOD samples, there is no causal relationship that indicates an increase in shape bias results in improved OOD robustness. Therefore, more focus should be on improving the OOD robustness directly, instead of trying to achieve higher robustness by aiming to increase a model's shape bias.

There is a dispute on whether there is a trade-off between in-domain accuracy and OOD robustness \cite{kurakin2016adversarial, herrmann2022pyramid}. If such a trade-off exists, focusing on improving model robustness to OOD examples comes at the cost of losing in-domain accuracy. By analyzing the results we obtain from studying the in-domain accuracy and OOD robustness of models trained on various augmented datasets, we find that both in-domain and OOD accuracies can be improved at the same time, and the trade-off does not always exist.

As we focus on improving model robustness to OOD samples, we should note that a very common threat to model robustness is the presence of bias in the dataset. Once identified, dataset bias can be partially reduced using data augmentation. ImageNet-1K \cite{deng2009imagenet} is known to be biased towards representing texture-related features \cite{geirhos2018imagenet}. By analyzing our results on the OOD robustness of models trained using augmented datasets, we find three sources of bias in the ImageNet-1K dataset that can be readily mitigated by applying data augmentations.

% In \cite{kurakin2016adversarial}, authors claim that there is a trade-off between in-domain and OOD accuracy. In \cite{herrmann2022pyramid}, however, it has been shown that this trade-off can be broken for ViT models, at least. 

% It should be noted that in this work, a metric for shape bias has also been introduced which is widely adopted by other researchers.

% Based on the hypothesis that an increase in shape bias results in an increase in OOD robustness, some researchers have focused on improving shape bias as their objective in order to achieve improved robustness. 

% We evaluate the OOD robustness of a diverse set of augmentations, including simulating weather conditions, occlusions, noise, color changes, etc., and study how each augmentation affects the model's robustness to OOD data. Based on our analysis, we find some biases in the ImageNet dataset that can be reduced using simple data augmentations. We show, however, that applying data augmentation does not always contribute to improving a model's OOD robustness. Following previous work, we investigate the influence of ``shape bias" on model robustness and experiment with different augmentations to show the relationship between in-domain accuracy, OOD accuracy, and shape bias. We also provide a strong basis to determine which augmentations to apply to the ImageNet dataset to achieve a higher OOD robustness. 
The main contributions of this work are the following:
\begin{itemize}

    \item We apply 39 data augmentations to ImageNet-100, a fair representative subset of ImageNet-1K, and evaluate the effect of each data augmentation on the OOD robustness of a ResNet-50 model.

    \item We find some biases in the ImageNet-1K dataset and suggest that by using simple data augmentations, most of these biases can be reduced. Moreover, we show that not all data augmentations improve the OOD robustness.

    \item We demonstrate that there is not necessarily a trade-off between in-domain and OOD robustness, and by choosing the right augmentations, both the in-domain and OOD accuracies can be improved.
    
    \item By investigating the relationship between shape bias and OOD robustness, we show that it is not good practice to focus on improving the model's shape bias in order to achieve higher OOD robustness. This is because an increase in a model's shape bias does not necessarily improve the model's robustness to OOD samples.

\end{itemize}

\section{Related Works}

In \cite{geirhos2018imagenet}, a stylized version of the ImageNet dataset, called Stylized ImageNet (SIN), has been created. Moreover, it is shown that a SIN-trained ResNet outperforms an ImageNet-trained ResNet on most of the image perturbations considered in their work. According to this result, it seems since SIN-trained ResNets learn a shape-based representation of ImageNet, they are more robust than ImageNet-trained ResNets that mostly focus on texture information. Therefore, the authors claim that ImageNet data induce texture bias in a network, and they hypothesize that by adding more shape bias to a network, model robustness will be improved.

This hypothesis forms the basis of some of the following related works in the literature. For example, in \cite{lee2022improving}, a shape-based augmentation method has been proposed that applies different augmentations from a set of three augmentations to the image foreground and background. The idea behind this is to reduce the texture bias and increase the shape bias to improve the model's robustness to OOD data. While this method has achieved acceptable results, it is limited to only three shape-based augmentations, including Color Jitter, Random Grayscale, and Random Gaussian Blur. Moreover, the questions of why these augmentations were selected and what makes them ``shape-based" remain unanswered.

Improving model robustness using shape-focused augmentations has also been observed in self-supervised learning methods. In \cite{hendrycks2019using} it is shown that using a rotation-predictor auxiliary head along with a classification network can improve model robustness since the model needs to learn the shape representations in order to minimize the loss function. 

Other than data augmentation, the choice of the network affects the shape bias, as well. Vision Transformers \cite{dosovitskiy2020image} are less biased towards texture compared to Convolutional Neural Networks (CNNs) \cite{naseer2021intriguing}. However, data augmentation has a much larger effect on shape bias compared to the network type or the training objective function \cite{geirhos2021partial, hermann2020origins}.

Although a correlation exists between using data augmentation and improving model robustness, it should not be concluded that such an increase in robustness is a direct result of selecting augmentations that impose shape bias on the model. In other words, \textit{an increase in shape bias results in an increase in model robustness} is only a hypothesis. To evaluate this hypothesis, in \cite{mummadi2021does}, a method is introduced to increase a model's shape bias by keeping the shape-related information in a training dataset while suppressing texture-related features. The authors challenged the hypothesis and concluded that there is no clear causal relationship between shape bias and model robustness. However, this method is limited because it only keeps the edge information in the input image, which cannot introduce meaningful shape information to the model when partial object occlusions exist in the image, for instance. Furthermore, they evaluated the OOD robustness of their models on only few image perturbations that are not diverse enough to represent a large variety of perturbations and do not include those that are commonly found in nature (except Frost).

% The work in \cite{geirhos2018imagenet} suggests that humans classify objects based on their shape, rather than texture. Therefore, if we want to make model decision-making more consistent with human decision-making, an increase in shape bias would be essential. Related to this, authors in \cite{tuli2021convolutional} claim that, unlike CNNs that inherently classify images based on local spatial features, attention-based models, such as ViTs, make decisions based on shape information rather than texture. Therefore, ViTs are more human-like in that aspect. However, \cite{geirhos2021partial} shows that even the state-of-the-art networks still suffer greatly from a lack of consistency with humans. More specifically, these models exhibit different error patterns compared to humans. Moreover, authors in \cite{hermann2020origins} show that even the networks that are designed to mimic human's visual system or those that replace convolutional layers with self-attention blocks do not have significantly lower texture biases compared with CNNs. Therefore, there is still much room for improvement in model-human consistency \cite{geirhos2021partial}. While according to these results, one might conclude that using ViTs is a better option compared to CNNs because ViTs are inherently less biased to texture \cite{naseer2021intriguing}, it has been shown that data augmentation has a much larger effect on shape bias compared to the network type or the training objective function \cite{geirhos2021partial, hermann2020origins}; leaving much more room for improvement in the augmentation side.

\section{Method}
We are interested in exploring the characteristics of different augmentations and investigating if there is a relationship between shape bias and a model's in-domain accuracy and OOD robustness. To this end, we performed an extensive set of experiments on a fairly representative subset of the ImageNet-1K dataset. The reason why we did not use ImageNet-1K was the significant computation requirements. More specifically, we needed to train a model for each augmented dataset with a training size two times larger than the original dataset. This required a huge amount of computation power if we wanted to use the whole ImageNet-1K dataset, especially in our case since we aimed to evaluate various augmentation types for the purpose of our studies. We used the experimental results to analyze biases in the ImageNet-1K dataset by examining those augmentations that affect the model OOD robustness the most. 

\subsection{Shape bias}
We calculate shape bias using a measure defined in \cite{geirhos2021partial} as described in \eqref{eq:shape_bias}. This measure is applied to a dataset containing images with a shape corresponding to one class, and a texture corresponding to another. For example, it contains an image with the texture of an elephant that shows the shape of a cat. If in this case, the model detects a cat, then, it has a stronger bias towards shape than texture. Therefore, in order to calculate shape bias using this method, we count the number of correct shape decisions and divide that by the number of correctly classified images (either shape or texture).

\begin{equation}
\label{eq:shape_bias}
\mbox{shape bias} = \frac{\mbox{\# correct shapes}}{\mbox{\# correct shapes + \# correct textures}}
\end{equation}

% B) Following \cite{islam2021shape}, we used an encoder $E$ to obtain shape dimensionality, denoted to as $|z_{shape}|$. We fed two images $I^a$ and $I^b$ of the same shape to the encoder and got latent representations $E(I^a) = z^a$ and $E(I^b) = z^b$, respectively. Then, we calculated the mutual information between these latent representations. We repeated the same process to calculate texture bias. Note that the image pairs we used were from the cue-conflict dataset. We used this\footnote{\url{https://github.com/islamamirul/shape_texture_neuron}} implementation to obtain shape bias values.

\subsection{OOD robustness}

To evaluate the OOD robustness of a model, first, we trained the model on the source dataset augmented using a selected augmentation type. Then, we evaluated the performance of the trained model on a target dataset with images that were obtained by applying 17 various perturbations to the images from the source dataset. We used the model's average top-1 classification accuracy on the target dataset as a measure to describe the OOD robustness in our experiments.

\section{Experiments}
\subsection{Dataset}
\subsubsection{OOD Benchmark Dataset}
 We used the OOD Benchmark dataset designed in \cite{geirhos2021partial} to evaluate the effect of each augmentation function on the OOD robustness. The OOD Benchmark dataset is a collection of 17 datasets that are obtained by applying different types of image perturbations on a subset of ImageNet-1K. Based on the perturbation type, these 17 datasets are categorized into two groups; five datasets belong to the Style group and the rest of the datasets belong to the Noise group. Each dataset in the Style group is created by applying a specific effect on a subset of the ImageNet-1K dataset. These effects include ``sketch", ``edge filter", ``silhouette", ``cue-conflict" (images that have a texture corresponding to a class with a shape from a different class), and ``stylized" (images that are stylized using painting styles). In the Noise group, images are obtained by applying different types of digital degradations to images, including noise and blur. Each image in these 17 datasets corresponds to a label from a set of 16 human-recognizable labels (``airplane", ``bear", ``bicycle", ``bird", ``boat", ``bottle", ``car", ``cat", ``chair", ``clock", ``dog", ``elephant", ``keyboard", ``knife", ``oven", ``truck") introduced in the same work. To obtain these labels, a mapping has been introduced from ImageNet-1K classes to 16 broader categories\footnote{\url{https://github.com/bethgelab/model-vs-human}}. For example, a few classes of different cat types, including ``tabby cat", ``Persian cat", and ``Siamese cat" are mapped to one class, called ``cat".

\subsubsection{ImageNet-100}
\label{sec:imagenet100}
For training purposes, we used a subset of 100 classes from the ImageNet-1K dataset, called ImageNet-100. In order to build ImageNet-100 from ImageNet-1K, we needed to pick 100 classes from the 1000 classes in ImageNet-1K. Since we wanted to perform OOD robustness testing on the OOD Benchmark dataset, we needed labels that could be converted to the 16 human-recognizable labels used in the OOD Benchmark dataset. Therefore, we examined the above mentioned mapping from ImageNet-1K to 16 human-recognizable categories to find those classes that were mapped to only few classes from ImageNet-1K. For example, ``knife" was only mapped to one class in ImageNet-1K, whereas ``dog" corresponded to 110 classes. To make ImageNet-100 balanced, we prioritized these classes over the ones that corresponded to a higher number of classes in ImageNet-1K. Then, according to the mapping, we started selecting ImageNet-1K classes until we reached 100 classes. This approach allowed us to end up with classes that can be mapped to the 16 human-recognizable labels, and at the same time, represent a balanced set of classes in ImageNet-1K. ImageNet-100 represents ImageNet-1K; because by design, for each class in ImageNet-1K, there is a broader class in ImageNet-100 that represents the selected ImageNet-1K class.

\subsubsection{Cue-conflict}
To calculate shape bias, we used the cue-conflict dataset which is one of the datasets in the Style group from the OOD Benchmark dataset. Cue-conflict is a collection of images that have the shape corresponding to class $A$, but texture corresponding to class $B$, where $A \neq B$.

\subsection{Implementation details}
We used the PyTorch framework for our experiments. For every augmentation, we trained a ResNet-50 model with an initial learning rate of 0.1 for 40 epochs and used the ReduceLROnPlateau scheduler that multiplies the learning rate by a factor of 0.1 whenever the validation accuracy has not increased for 5 epochs. The batch size and weight decay were set to 256 and 0.0005, respectively. We also trained a baseline model, called Vanilla, that is trained on the ImageNet-100 training set without adding any augmentations for comparison purposes.
For each of the 39 models, one for each augmentation type, we evaluated the in-domain accuracy on clean data, as well as the model's OOD robustness when evaluated on the OOD Benchmark dataset. Furthermore, we used the cue-conflict dataset from the OOD Benchmark dataset to measure the shape bias of each model.

\subsection{Augmentations}

\begin{figure*}[h!]
\centering
\includegraphics[width=\textwidth]{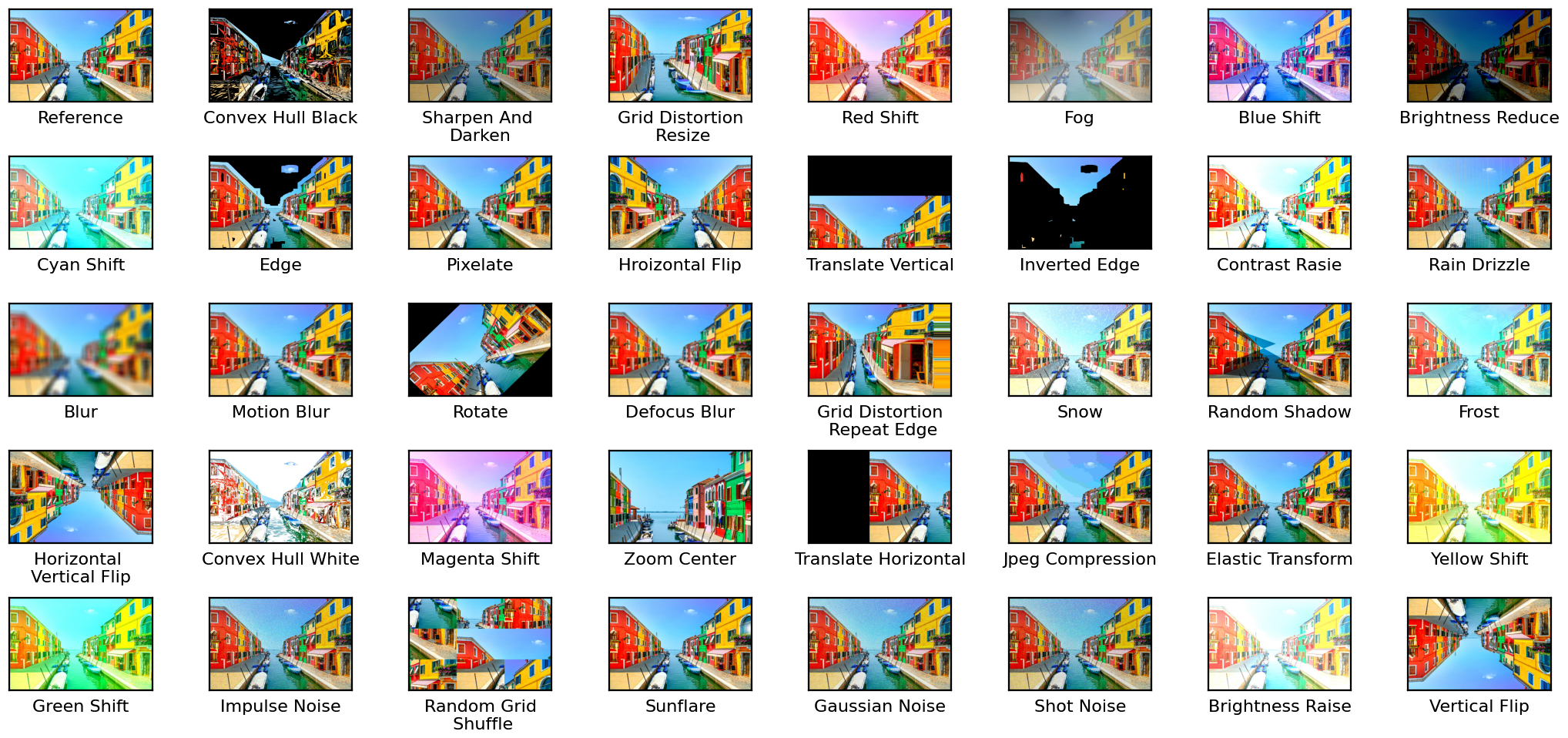}
\caption{An overview of how different augmentations alter the reference (original) image. Best viewed in color.}
\label{fig:augmentation}
\end{figure*}

We studied 39 types of augmentations that span a diverse set of image perturbation types. The intensity of these augmentations is balanced in a way that the transformed image is still easily recognizable to an observer, while they can instantly realize that some known kind of perturbation has been applied to the image. Fig. \ref{fig:augmentation} illustrates different types of augmentations and how they change a reference image.

% There are four composed augmentations studied as well, namely, Texture, Shape, Mixed Texture Shape, and Mixed Shape Texture. Each of these composed augmentations is an ordered list of three augmentations that are chosen based on their effect on a fixed ResNet-50 model's shape and texture bias. To decide which augmentations to choose for each list, we evaluated the shape bias of each model trained on an augmented dataset and sorted the augmentations based on their effect on the model's shape bias. Then, each list was composed as follows:
% - Shape: Blur, Frost, Impulse Noise (Top three augmentations with the highest effect on shape bias)
% - Texture: Vertical Flip, Grid Distortion Repeat Edge, Rotate (top three augmentations with the lowest effect on shape bias)
% - Mixed Shape Texture: Blur, Contrast Raise, Vertical Flip (top augmentation with the highest effect on shape bias, a selected augmentation with a medium shape bias, top augmentation with the lowest effect on shape bias)
% - Mixed Texture Shape: Vertical Flip, Contrast Raise, Blur (top augmentation with the lowest effect on shape bias, a selected augmentation with a medium shape bias, top augmentation with the highest effect on shape bias)

% The order comes into play during the training. We divide the number of epochs by three and apply each augmentation during the first, second, or third 1/3 of epochs. This way, we can see the effect of using texture-based or shape-based augmentations first.

\section{Results and Discussion}
\subsection{Evaluating augmentations on the OOD Benchmark}
\begin{figure*}[h!]
\centering
\includegraphics[width=\textwidth]{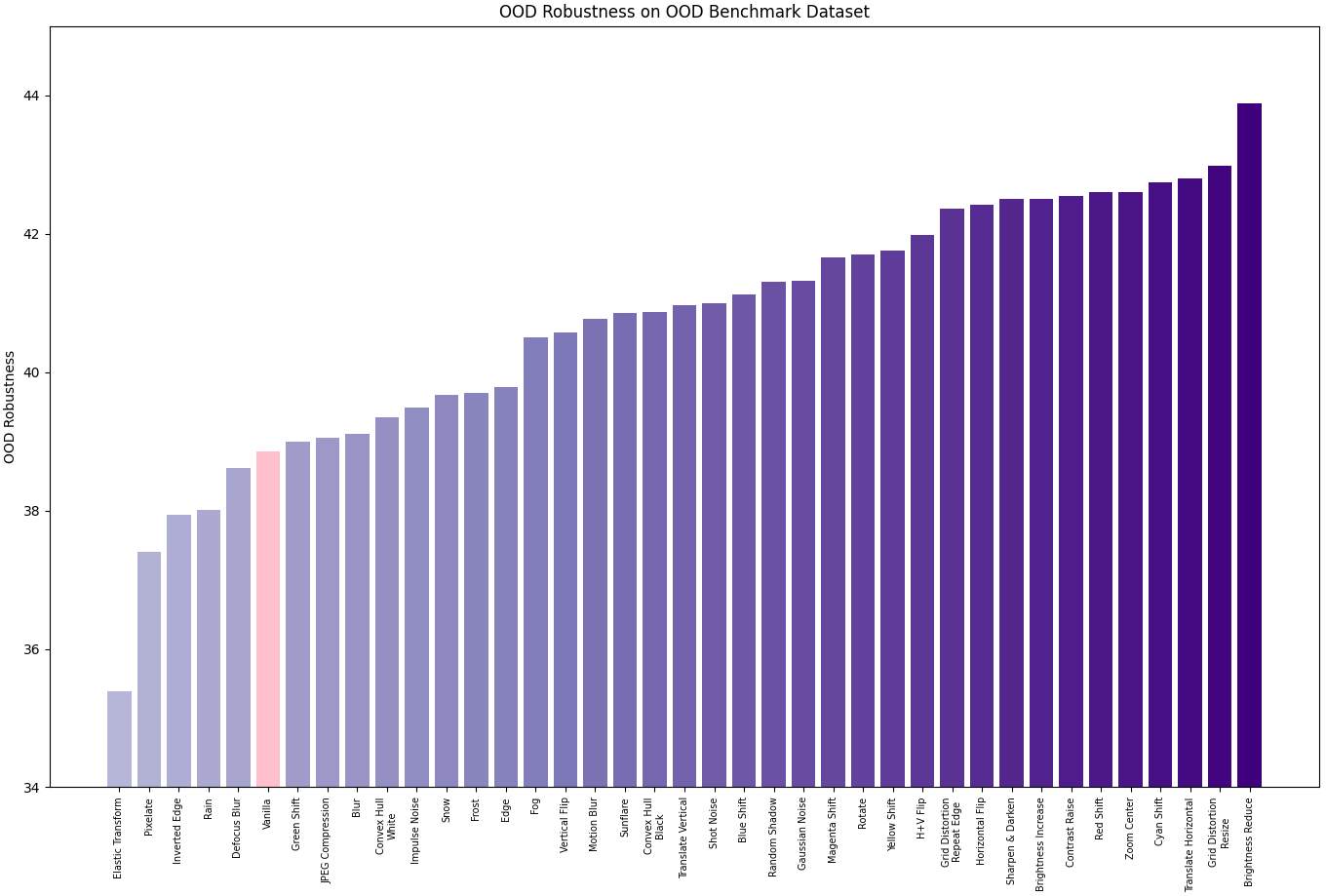}
\caption{Evaluation of OOD robustness for various augmentations on the OOD Benchmark dataset. The pink bar represents the Vanilla (baseline with no data augmentations) model.}
\label{fig:sorted_ood_on_ood}
\end{figure*}

By evaluating ResNet-50 models trained on augmented datasets using various image augmentations, we calculated the OOD robustness on the OOD Benchmark dataset. We measured this factor by considering a diverse set of 39 augmentations across various perturbation groups, including blur, noise, distortion, occlusion, color shift, lightness, weather, and geometry. We also measured the OOD robustness for the Vanilla (baseline) model, which is trained on ImageNet-100 without applying any data augmentation to compare the results. Fig. \ref{fig:sorted_ood_on_ood} shows how OOD robustness varies across different augmentations.

As we mentioned in Section \ref{sec:imagenet100}, ImageNet-100 fairly represents ImageNet-1K by design. Therefore, by choosing the augmentations that have the highest effect on model robustness according to Fig. \ref{fig:sorted_ood_on_ood} and augmenting ImageNet-1K using those augmentations, we can improve model robustness in situations where we are using the ImageNet-1K dataset. Fig. \ref{fig:sorted_ood_on_ood} illustrates that ``Brightness Reduce" augmentation contributed the most to OOD robustness. On the other hand, ``Brightness Increase" holds the 8th rank out of 40. This means that most of the images are in the same range of light settings in the ImageNet-1K dataset and a strong light augmentation can increase OOD robustness significantly.

The second most effective augmentation is ``Grid Distortion Resize" which performs grid distortion on the input image and then resizes the output to the original image size. This augmentation alters the visual representation of the image in a way that objects seem more narrow or wide on different sides. This is similar to viewing an object from different angles. Since applying this augmentation has a high effect on model robustness, it can be concluded that the ImageNet-1K dataset images mostly represent objects from limited angles. To improve OOD robustness against other points of view, one cost-efficient approach is to use geometric augmentations, like the ``Grid Distortion Resize" augmentation.

Last but not least, ``Translate Horizontal" stands third in the ranking, which indicates that ImageNet-1K objects are usually located in the center of the image; whereas in real-world applications, this is not always the case. By applying a simple translate augmentation, OOD robustness can improve significantly.

\subsection{Does data augmentation always improve OOD robustness?}
According to Fig. \ref{fig:sorted_ood_on_ood}, some of the augmentations rank lower than the baseline Vanilla model. For instance, ``Elastic Transform" which adds a wavy effect on the input image has a significantly lower OOD robustness (-3.5\%) compared to the Vanilla model. This observation is an example of a case where data augmentation does not improve model robustness. Since such cases exist, we should always perform OOD robustness evaluations on augmented datasets to make sure we are achieving results that are better than the baseline.

\subsection{Is there a trade-off between in-domain accuracy and OOD robustness?}
In \cite{kurakin2016adversarial}, it is claimed that there is a trade-off between in-domain accuracy and OOD robustness. In \cite{herrmann2022pyramid}, however, it has been shown that this trade-off can be broken for Vision Transformers, at least. We study if such a relationship exists for CNN-based models by exploiting the in-domain accuracy and OOD robustness of our trained models on the OOD Benchmark dataset.
Table \ref{tab:tradeoff}, demonstrates a few examples for which this trade-off breaks. A few other augmentations meet these conditions as well, which are not listed in this table for brevity. According to these examples, it cannot be concluded that increasing the OOD robustness comes at the cost of a decrease in in-domain accuracy, necessarily. An ideal augmentation increases both in-domain accuracy and OOD accuracy.

% \begin{figure*}[h!]
% \centering
% \includegraphics[width=\textwidth]{id vs ood.png}
% \caption{This figure shows the OOD robustness of augmentations sorted by their in-domain accuracy. The vertical and horizontal lines represent the Vanilla (baseline) model's in-domain accuracy and OOD robustness, respectively.}
% \label{fig:appenix_tradeoff_breaks}
% \end{figure*}

\begin{table}[h!]
\caption{This table shows a few augmentation examples that can be applied to the ImageNet-100 dataset and increase both in-domain accuracy and OOD robustness, simultaneously. Avg. OOD Robustness is the weighted average of OOD-Noise and OOD-Style robustness values.}
\label{tab:tradeoff}
\centering
\resizebox{\columnwidth}{!}{%
\begin{tabular}{ccccc}
\toprule
     \begin{tabular}[c]{@{}c@{}} Augmentation \\ Name
     \end{tabular} &
    \begin{tabular}[c]{@{}c@{}} In-domain \\ Accuracy
    \end{tabular} &
    \begin{tabular}[c]{@{}c@{}}
     OOD-Noise \\ Robustness
     \end{tabular} &
    \begin{tabular}[c]{@{}c@{}} OOD-Style \\ Robustness 
    \end{tabular} &
    \begin{tabular}[c]{@{}c@{}} Avg. OOD \\ Robustness
    \end{tabular}\\
\midrule
Vanilla (Baseline) & 89.06 & 42.99 & 28.92 & 38.85\\
\midrule
Grid Distortion Resize & 94.92 & 48.67 & 29.33 & 42.98\\
\midrule
Grid Distortion Repeat Edge & 93.75 & 47.47 & 30.11 & 42.3\\
\midrule
Rotate & 93.75 & 46.27 & 30.74 & 41.70\\
\midrule
Translate Vertical & 92.57 & 44.44 & 32.64 & 40.96\\
\midrule
Translate Horizontal & 92.18 & 46.8 & 33.19 & 42.79\\
\midrule
Zoom Center & 91.01 & 47.16 & 31.66 & 42.60\\
\bottomrule
\end{tabular} %
}
\end{table}

\subsection{The relationship between shape bias and OOD robustness}
\begin{figure}[h!]
\centering
\includegraphics[width=\columnwidth]{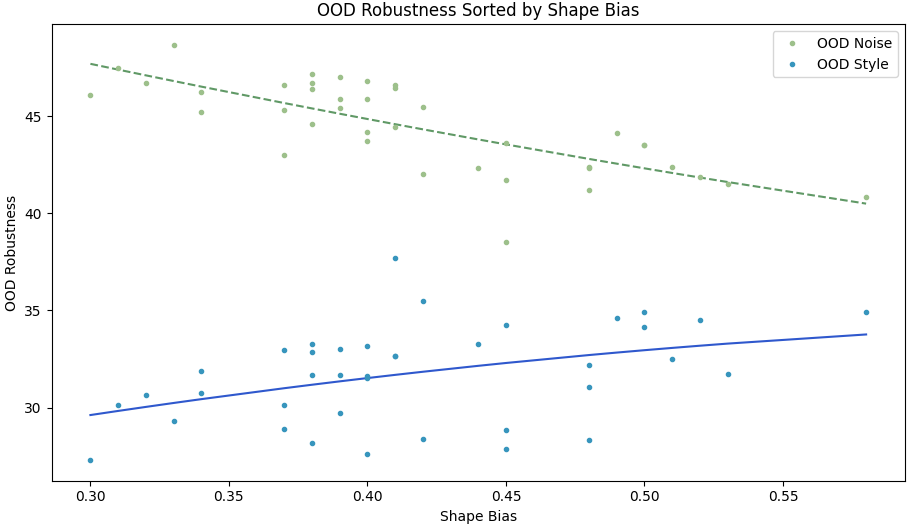}
\caption{This figure shows the relationship between the shape bias and the OOD robustness of models trained on augmented datasets. The evaluation has been performed on the OOD Benchmark dataset that contains two groups of datasets, called the Noise and the Style groups.}
\label{fig:shape_bias_vs_ood}
\end{figure}

An increase in the shape bias does not \textit{result} in an increase in the OOD robustness. However, as the shape bias increases, OOD robustness against the augmentations in the Style group \textit{usually} increases at the cost of a decrease in the OOD robustness against the augmentations in the Noise group. This claim can be observed in Fig. \ref{fig:shape_bias_vs_ood}, where the data points are estimated using a second-degree polynomial for each OOD group (Style and Noise). Since the OOD robustness for the Noise group reduces with the increase in shape bias, we cannot claim that a higher shape bias results in a higher OOD robustness. Therefore, it is not good practice to try to achieve improved model robustness by focusing on improving the model's shape bias as seen in the literature. 

\section{Conclusion}
% The conclusion goes here. this is more of the conclusion
In this work, we evaluated the OOD robustness of 39 different types of augmentations on the OOD Benchmark dataset using ResNet-50 models trained on various augmented versions of the ImageNet-100 dataset. By studying the experimental results carefully, we found some biases in the ImageNet-1K dataset that can be reduced by applying data augmentation. Moreover, we explained that not all data augmentations improve model robustness, and it is necessary to evaluate whether augmenting a dataset using a specific type of data augmentation contributes to an increase in OOD robustness. Furthermore, the relationship between in-domain accuracy and OOD robustness was studied, and it was shown that there is not necessarily a trade-off between in-domain accuracy and OOD robustness. Finally, it was concluded that an increase in shape bias does not always result in an increase in OOD robustness.

% \section*{Acknowledgment}

% more thanks here



%\bibliographystyle{IEEEtran}

\bibliographystyle{IEEEtran}
\bibliography{main}

\end{document}